\documentclass[runningheads]{llncs}
\usepackage[T1]{fontenc}
\usepackage{graphicx}
\def \ie{{\em i.e.,}}

\def \one{{\em i)}}
\def \two{{\em ii)}}

\def \Robust{\mathrm{R}}

\usepackage{hyperref}
\usepackage{ amssymb,amsmath, mathtools}
\usepackage{float}
\DeclarePairedDelimiter{\norm}{\lVert}{\rVert} 
\usepackage{amsfonts}
\usepackage{pifont}
\usepackage{multirow}
\usepackage{rotating}
\usepackage{graphicx}
\usepackage{algorithm}
\usepackage{xcolor}
\usepackage{algorithmic}
\usepackage{makecell}

\newcommand{\cmark}{\ding{51}}%
\newcommand{\xmark}{\ding{55}}
\usepackage{booktabs}

\newcommand{\obj}{o}
\newcommand{\of}{w}

\usepackage{comment}

\begin{document}
\title{Model-Agnostic Reachability Analysis on Deep Neural Networks}
%
%
%
\author{Chi Zhang\inst{1} \and
Wenjie Ruan\inst{1}$^{\dagger}$\and
Fu Wang\inst{1} \and
\\Peipei Xu\inst{2} \and
Geyong Min\inst{1} \and
Xiaowei Huang\inst{2}}
\authorrunning{C. Zhang, W. Ruan et al.}
%
\institute{Department of Computer Science, University of Exeter, Exeter, UK \and
Department of Computer Science, University of Liverpool, Liverpool, UK\\
\{cz338; w.ruan; fw377; g.min\}@exeter.ac.uk; \\ \{peipei.xu; xiaowei.huang\}@liverpool.ac.uk;~~~$\dagger~$Corresponding Author}

\maketitle              

\begin{abstract}
Verification plays an essential role in the formal analysis of safety-critical systems. Most current verification methods have specific requirements when working on Deep Neural Networks (DNNs). They either target one particular network category, e.g., Feedforward Neural Networks (FNNs), or networks with specific activation functions, e.g., ReLU. In this paper, we develop a model-agnostic verification framework, called DeepAgn, and show that it can be applied to FNNs, Recurrent Neural Networks (RNNs), or a mixture of both. Under the assumption of Lipschitz continuity, DeepAgn analyses the reachability of DNNs based on a novel optimisation scheme with a global convergence guarantee. It does not require access to the network’s internal structures, such as layers and parameters. Through reachability analysis, DeepAgn can tackle several well-known robustness problems, including computing the maximum safe radius for a given input,
and generating the ground-truth adversarial example. 
We also empirically demonstrate DeepAgn's superior capability and efficiency in handling a broader class of deep neural networks, including both FNNs and RNNs with very deep layers and millions of neurons, than other state-of-the-art verification approaches. Our tool is available at \url{https://github.com/TrustAI/DeepAgn} 
\end{abstract}

\vspace{-0.8cm}
\section{Introduction}

DNNs, or systems with neural network components, are widely applied in many applications such as image processing, speech recognition, and medical diagnosis \cite{goodfellow2016}. However, DNNs are vulnerable to adversarial examples~\cite{szegedy2013intriguing} \cite{huang2020survey} \cite{yin2022dimba}. It is vital to analyse the safety and robustness of DNNs before deploying them in practice, particularly in safety-critical applications. 

The research on evaluating the robustness of DNNs mainly falls into two categories: falsification-based and verification-based approaches. 
While falsification approaches (e.g. adversarial attacks)~\cite{goodfellow2014explaining}
can effectively find adversarial examples, they cannot provide theoretical guarantees.
Verification techniques, on the other hand, can rigorously prove the robustness of deep learning systems with guarantees \cite{mu20223dverifier,mu2023certified,huang2020survey,huang2017safety,ruan2018reachability}. 
Some researchers propose to reduce the safety verification problems to constraint satisfaction problems that can be tackled by constraint solvers such as Mixed-Integer Linear Programming (MILP)~\cite{akintunde2018reachability},
Boolean Satisfiability (SAT)~\cite{narodytska2018verifying}, or Satisfiability Modulo Theories (SMT)~\cite{katz2017reluplex}. Another popular technique is to apply search algorithms~\cite{huang2017safety} or Monte Carlo tree search~\cite{WWRHK2018} over discretised vector spaces on the inputs of DNNs.  
To improve the efficiency, these methods can also be combined with a heuristic searching strategy to search for a counter-example or 
an activation pattern that satisfies 
certain constraints, such as SHERLOCK~\cite{dutta2017output} and Reluplex~\cite{katz2017reluplex}.
Nevertheless, the study subjects of these verification methods are restricted. They either target specific layers (e.g., fully-connected or convolutional layers), have restrictions on activation functions (e.g., ReLU activation only), or are only workable on a specific neural network structure (e.g., feedforward neural networks).
Particularly, in comparison to FNNs, verification on RNNs is still in its infancy, with only a handful of representative works available, including \cite{ko2019popqorn,jacoby2020verifying,zhang2020verification}. The adoption of \cite{ko2019popqorn} requires short input sequences, and \cite{jacoby2020verifying,zhang2020verification} can result in irresolvable over-approximation error.
\begin{figure*}[t]
\centering
\includegraphics[scale=0.35]{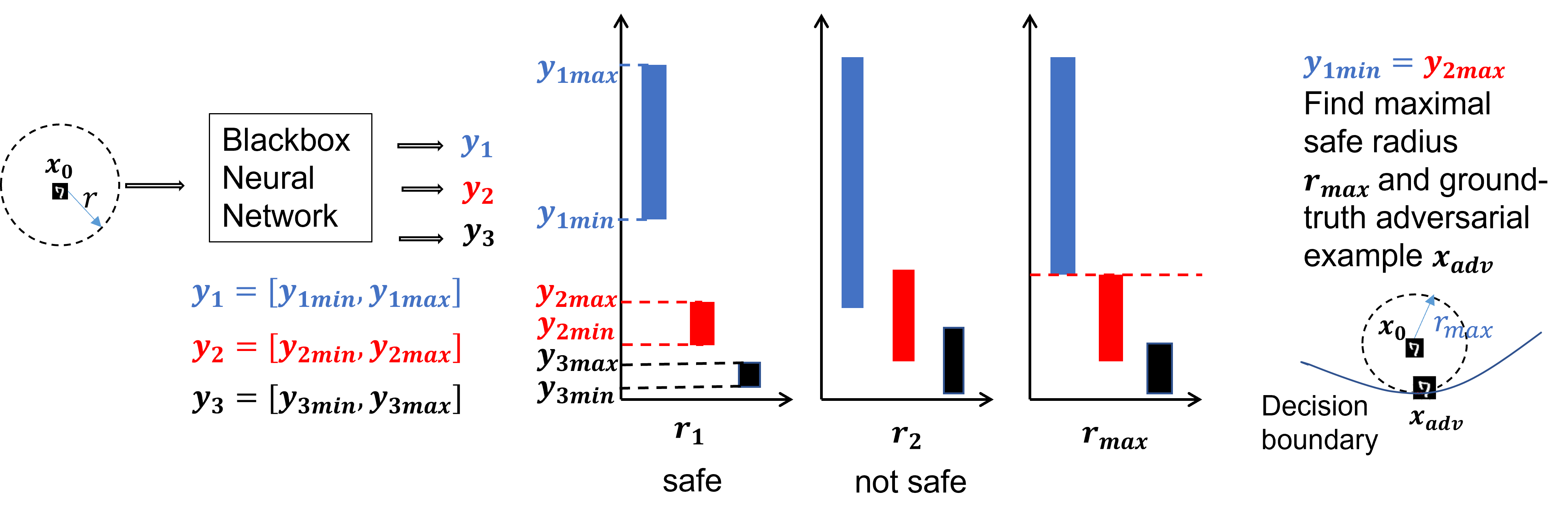}
\vspace{-6mm}
\caption{Illustration of DeepAgn working on a black-box three-output neural network. {\em In reachability problem, given a set of inputs (quantified by a predefined $L_p$-norm ball) and a well-trained black-box neural network, DeepAgn can calculate the output range, namely, the minimal and maximum output confidence of each label (i.e., $[y_{1min},y_{1max}]$, $[y_{2min},y_{2max}]$, and $[y_{3min},y_{3max}]$). For the safety verification problem, we can use a binary search upon the reachability to find the maximum safe radius $r_{max}$ where the confidence intervals of the original label $y_1$ and target label $y_2$ meet.}
}
\label{fig:overall}
\vspace{-5mm}
\end{figure*}

This paper proposes a novel model-agnostic solution for safety verification on both feedforward and recurrent neural networks without suffering from the above weaknesses. Figure~\ref{fig:overall} outlines the working principle of DeepAgn, demonstrating its safety evaluation process and the calculation of the maximum safety radius. 
To the best of our knowledge, DeepAgn is one of the pioneering attempts on {\em model-agnostic} verification that can work on both modern feedforward and recurrent neural networks under a unified framework. DeepAgn can deal with DNNs with very deep layers, a large number of neurons, and any type of activation function, via a black-box manner (without access to the internal structures/parameters of the network). Our contributions are summarised below:
\begin{itemize}
    \item To theoretically justify the applicability of DeepAgn, we prove that recurrent neural networks are also Lipschitz continuous for bounded inputs.
    
    \item We develop an efficient method for reachability analysis on DNNs. We demonstrate that this {\em generic} and {\em unified} model-agnostic verification framework can work on FNNs, RNNs, and a hybrid of both. DeepAgn is an anytime algorithm, i.e., it can return both intermediate lower and upper bounds that are gradually, but strictly, improved as the computation proceeds; and it has provable guarantees, i.e., both the bounds can converge to the optimal value within an arbitrarily small error with provable guarantees.
    
    \item Our experiments demonstrate that DeepAgn outperforms the state-of-the-art verification tools in terms of both accuracy and efficiency when dealing with complex, large and hybrid deep learning models.
\end{itemize}

\vspace{-0.3cm}
\section{Related Work}
\label{sec:related work}
\vspace{-0.3cm}


\textbf{Adversarial Attacks~~} Attacks apply heuristic search algorithms to find adversarial examples. Attacking methods are mainly guided by the forward or cost gradient of the target DNNs. Major approaches include L-BFGS~\cite{szegedy2013intriguing}, FGSM~\cite{goodfellow2014explaining}, Carlini \& Wagner attack~\cite{carlini2017towards}, Universal Adversarial Attack~\cite{zhang2023generalizing}, etc.
Adversarial attacks for FNNs can be applied to cultivate adversarial examples for RNNs with proper adjustments. The concepts of adversarial example and adversarial sequence for RNNs are introduced in~\cite{papernot2016crafting}, in which they concrete adversarial examples for Long Short Term Memory (LSTM) networks. Based on the C\&W attack~\cite{carlini2017towards}, attacks are implemented against DeepSpeech in~\cite{carlini2018audio}. The method in~\cite{gong2017crafting} is the first approach to analyse and perturb the raw waveform of audio directly.
\vspace{2mm}


\begin{table*}[t]\small
 
  \caption{Comparison with other verification techniques from different aspects}
  \label{tab-comparison}
  \scalebox{0.57}
  {
  \begin{tabular}{l|ccc ccc ccc}
     \toprule
     & \multicolumn{1}{|c}{\vtop{\hbox{\strut \textbf{Guarantees}} }}
         &{\vtop{\hbox{\strut \textbf{Core Techniques} }}}
         &{\vtop{\hbox{\strut  \textbf{Neural Network Types}  }}} 
         &{\vtop{\hbox{\strut  \textbf{\makecell{Model\\Agnostic}} }}} 
         &{\vtop{\hbox{\strut  \textbf{\makecell{Exact\\Computation}} }}} 
         &{\vtop{\hbox{\strut  \textbf{Model Access} }}} \\ \hline 
        \textbf{Reluplex}~\cite{katz2017reluplex} &  {Deterministic} & {SMT+LP} & {ReLu-based FNNs}  &{ \xmark} &{ \cmark} &{Model parameters}\\
        \hline
        \textbf{Planet}~\cite{ehlers2017formal} &  {Deterministic  }& { SAT+LP } & { ReLU-based FNNs} &{ \xmark}  &{ \cmark}&{Model parameters} \\
        \hline
        \textbf{AI2}~\cite{gehr2018ai2}&  {Upper bound}& {Abstract Interpretation} & {ReLU-based FNNs} &{ \xmark}  &{ \xmark} &{Model Parameters} \\
        \hline
        \textbf{ConDual}~\cite{wong2018provable} &  { Upper bound }& {Convex relaxation }& {ReLU-based FNNs} &{ \xmark}  &{ \xmark}&{Model parameters}\\ 
        \hline
        \textbf{DeepGO}~\cite{ruan2018reachability}& {Converging bound} & {Lipschitz Optimisation } & \makecell{FNNs with Lipschitz continuous \\layers (ReLU, Sigmoid, Tanh, etc.)} &{ \xmark}  & {\cmark}&{Confidence values}\\ 
        \hline
        \textbf{FastLip}~\cite{weng2018towards} & { Upper bound} & {Lipschitz estimation} & {ReLU-based FNNs}  &{ \xmark}   &{ \xmark}&{Model parameters}\\
        \hline
        \textbf{DeepGame}~\cite{WWRHK2018} & \makecell{Approximated\\converging bound} & {Search based} & \makecell{ReLU/Tanh/Sigmod based FNNs}    &{ \xmark}  &{ \cmark}&{Confidence values}\\ 
        \hline
        \textbf{POPQORN} ~\cite{ko2019popqorn} & {Upper bound} & {Unrolling } & {RNNs, LSTMs, GRUs}  &{ \xmark}    &{ \xmark}&{Model parameters}  \\
        \hline
        \textbf{RnnVerify} ~\cite{jacoby2020verifying} & {Upper bound} & {Invariant Inference} & { RNNs }  &{ \xmark}    &{ \xmark}&{Model parameters} \\
        \hline
        \textbf{VERRNN} ~\cite{zhang2020verification} & {Upper bound} & {Unrolling+MILP} & {RNNs}  &{ \xmark}   &{ \xmark}&{Model parameters}   \\
        \hline
        \textbf{DeepAgn}  & { Converging bound} & {Lipschitz Optimisation} & \makecell{FNNs (CNNs), RNNs, Hybrid networks \\ with Lipschitz continuous layers}  &{ \cmark} & {\cmark} &{Confidence values}\\
     \bottomrule
  \end{tabular} }
  \vspace{-5mm}
\end{table*}

\noindent\textbf{Verification on DNNs~~} The recent advances of DNN verification include a layer-by-layer exhaustive search approach~\cite{huang2017safety}, methods using constraint solvers~\cite{PT2010} \cite{katz2017reluplex}, global optimisation approaches~\cite{ruan2018reachability} \cite{wang2023towards}, and the abstract interpretation approach~\cite{gehr2018ai2} \cite{mirman2018differentiable}. 
The properties studied include robustness~\cite{huang2017safety,katz2017reluplex,zhang2022proa}, or reachability~\cite{ruan2018reachability}, i.e., whether a given output is possible from properties expressible with SMT constraints, or a given output is reachable from a given subspace of inputs. Verification approaches aim to not only find adversarial examples but also provide guarantees on the results obtained. However, efficient verification on large-scale deep neural networks is still an {\em open problem}. Constraint-based approaches such as Reluplex can only work with a neural network with a few hundred hidden neurons~\cite{PT2010,katz2017reluplex}. Exhaustive search suffers from the state-space explosion problem~\cite{huang2017safety}, although it can be partially alleviated by Monte Carlo tree search~\cite{WWRHK2018}. Moreover, the work~\cite{akintunde2018reachability} considers determining whether an output value of a DNN is reachable from a given input subspace. It proposes a MILP-based solution. SHERLORCK~\cite{dutta2017output} studies the range of output values from a given input subspace. This method interleaves local search (based on gradient descent) with global search (based on reduction to MILP). Both approaches can only work with small neural networks.

The research on RNN verification is still relatively new and limited compared with verification on FNNs. 
Approaches in ~\cite{ko2019popqorn,zhang2020verification,vengertsev2020recurrent} start with unrolling RNNs and then use the equivalent FNNs for further analysis.
POPQORN~\cite{ko2019popqorn} is an algorithm to quantify the robustness of RNNs, in which upper and lower planes are introduced to bound the non-linear parts of the estimated neural networks. 
The authors in~\cite{jacoby2020verifying} introduce invariant inference and over-approximation, transferring the RNN to a simple FNN model, demonstrating better scalability. However, the search for a proper invariant form is not straightforward. 
In Table~\ref{tab-comparison}, we compare DeepAgn with other safety verification works from {\em six} aspects. DeepAgn is the {\em only} model-agnostic verification tool that can verify hybrid networks consisting of both RNN and FNN structures. DeepAgn only requires access to the confidence values of the target model, enabling the verification in a {\em black-box} manner. Its precision can reach an arbitrarily small (pre-defined) error with a global convergence {\em guarantee}.
\vspace{-0.5cm}
\section{Preliminaries}
\vspace{-0.3cm}
Let $\obj: [0,1]^m \rightarrow \mathbb{R}$ be a generic function that is Lipschitz continuous. The generic term $\obj$ is cascaded with the Softmax layer of the neural network for statistically evaluating the outputs of the network. Our problem is to find its upper and lower bounds given the set $X'$ of inputs to the network.


\begin{definition}[Generic Reachability of Neural Networks]
Let $X'\subseteq [0,1]^n$ be an input subspace and $f: \mathbb{R}^n \rightarrow \mathbb{R}^m$ is a neural network. The generic reachability of neural networks is defined as the reachable set $R(\obj,X',\epsilon) = [l,u]$ of network $f$ over the generic term $\obj$ under an error tolerance $\epsilon \geq 0$ such that 
\vspace{-0.2cm}
\begin{equation}\label{eqn:pr-1}
\begin{split}
   \inf_{x' \in X'} \obj(f(x')) -\epsilon  \leq l \leq  \inf_{x' \in X'} \obj(f(x'))  + \epsilon \\
 \ \sup_{x' \in X'} \obj(f(x')) - \epsilon \leq  u \leq \sup_{x' \in X'} \obj(f(x')) + \epsilon
 \end{split}
\end{equation}

We write $u(\obj,X',\epsilon)=u$ and $l(\obj,X',\epsilon)=l$ for the upper and lower bound respectively. Then the reachability diameter is 
	\label{eqn:pr-2}
   $        D(\obj,X',\epsilon) =u(\obj,X',\epsilon)-l(\obj,X',\epsilon)$
Assuming these notations, we may write $D(\obj,X',\epsilon;f)$ if we need to explicitly refer to the network $f$. 
\end{definition}

\begin{definition}[Safety of Neural Network]
A network $f$ is safe with respect to an input $x_0$ and an input subspace $X'\subseteq [0,1]^n$ with $x_0 \in X'$, 
if 
\vspace{-0.2cm}
\begin{equation}
\forall x' \in X': \arg\max_{j} c_j(x') = \arg\max_{j} c_j(x_0)
\vspace{-0.3cm}
\end{equation} 
where $c_j(x_0) = f(x_0)_j$ returns $N$'s confidence in classifying $x_0$ as label $j$. 
\end{definition}

\begin{definition}[Verified Safe Radius] Given a neural network $f: \mathbb{R}^n \rightarrow \mathbb{R}^m$  and an input sample $x_0$, a verifier $V$ returns a verified safe radius $r_v$  regarding the safety of neural network. For input $x'$ with $\|x'-x_0\|\le r_v$, the verifier guarantees that $\arg\max_{j} c_j(x') = \arg\max_{j} c_j(x)$. For $\|x'-x_0\|>r$, the verifier either 
confirms $\arg\max_{j} c_j(x') \neq \arg\max_{j} c_j(x)$ or provides an unclear answer.
\end{definition}

\begin{figure}[H]
\vspace{-0.5cm}
	\centering
	\includegraphics[width=0.5\linewidth]{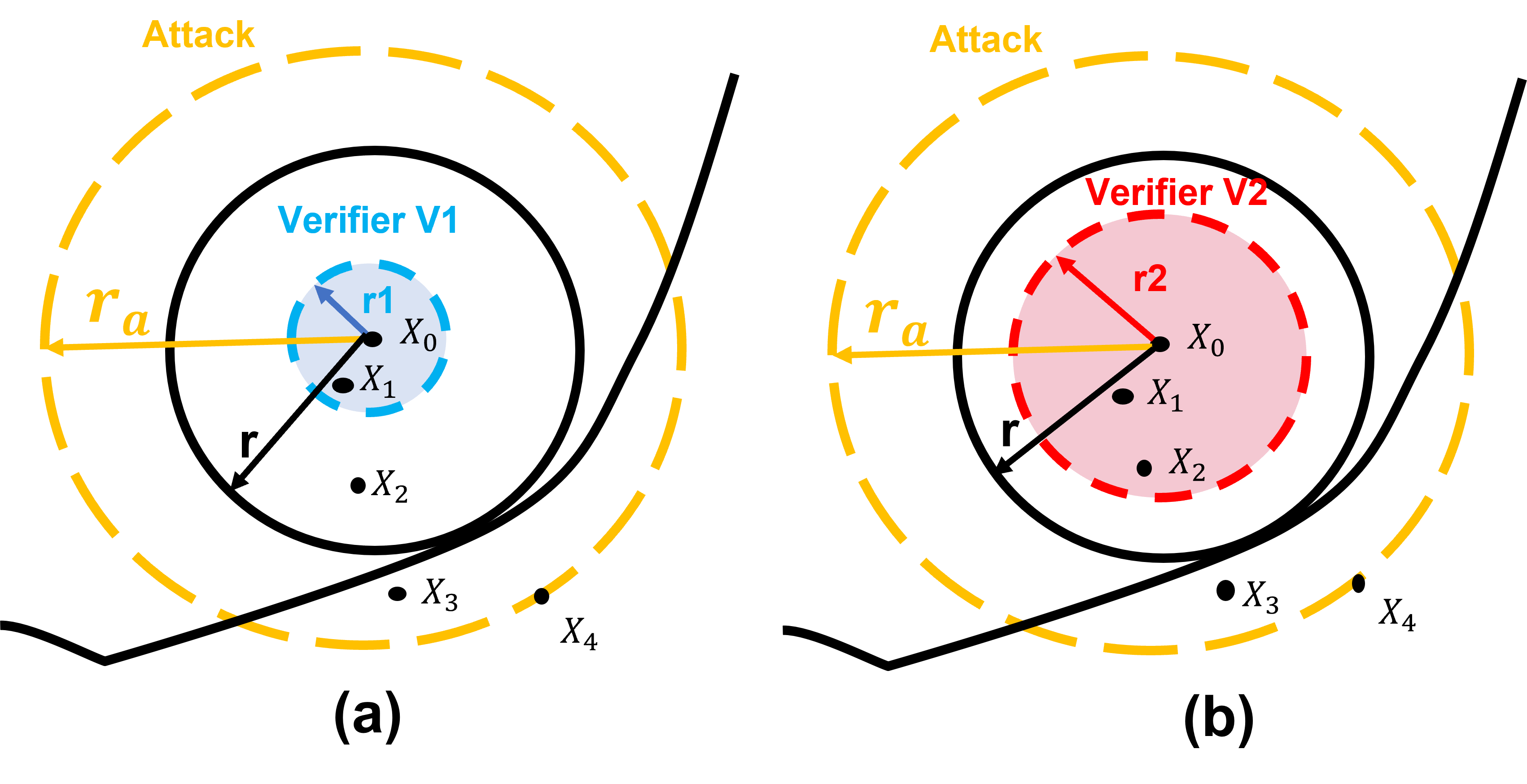}
 \vspace{-0.5cm}
	\caption{Verification of samples $x_1, x_2, x_3, x_4$ by different verifiers (a) Verifier V1 (b) Verifier V2, with verified safe radius $r_1<r_2$.  {\em According to V1, $x_1$ is safe since  $\|x_1-x_0\|<r_1$. $x_2, x_3$ and $x_4$ are determined as unsafe since $\|x_2-x_0\|>r_1,\|x_3-x_0\|>r_1,\|x_4-x_0\|>r_1$.  According to V2, $x_1,x_2$ are safe since $\|x_1-x_0\|<r_2,\|x_2-x_0\|<r_2$. $ x_3, x_4$ are determined as unsafe since  $\|x_3-x_0\|>r_2,\|x_4-x_0\|>r_2$. $x_4$ is generated by the attack method. Adversarial example $x_3$ still exists in the attack radius.} }
	\label{fig-1}
\end{figure}

Verified safe radius is important merit for robustness analysis, which is adopted by many verification tools such as CLEVER\cite{weng2018evaluating} and POPQORN \cite{ko2019popqorn}. Verification tools can further determine the safety of the neural network by comparing the verified safe radius $r_v$ and the perturbation radius. A neural network $f$ is determined safe by verifier $V$ with respect to input $x_0$, if $ \|x'-x_0\|\le r_v$. 
In Figure \ref{fig-1}, the verification tool $V_2$ with higher verified $r_2>r_1$ radius have a higher evaluation accuracy. The sample $x_2$ is misjudged as unsafe by $V_2$.

\vspace{-0.8cm}

 \vspace{-0.3cm}
\begin{definition}[Maximum Radius of a Safe Norm Ball]
Given a  neural network  $f:\mathbb{R}^{m\times{n}}\to\mathbb{R}^s$, an distance metric $\|\centerdot\|_D$, an input $x_0\in\mathbb{R}^{m\times{n}}$, a norm ball $B(f,x_0,\|\centerdot\|_D,r)$ is a subspace of $[a,b]^{m\times{n}}$ such that $B(f,x_0,\|\centerdot\|_D,r)=\{x'|\|x'-x_0\|_D\le{r}\}$. 
When $f$ is safe in $B(f,x_0,\|\centerdot\|_D,r)$ and not safe in any input subspace $B(f,x_0,\|\centerdot\|_D,r')$ with $r'>r$, we call $r$ here the maximum radius of a safe norm ball.
\end{definition}
 \vspace{-0.3cm}
\begin{definition}[Successful Attack on Inputs]
Given a neural network $f$ and input $x_0$, a $\alpha$-bounded attack $A_{\alpha}$ create input sets $X'=\{x', \|x'-x_0 \| \le \alpha \}$. $A_{\alpha}$ is a successful attack, if an $x_a\in X'$ exists, where $ \arg\max_{j} c_j(x_a) \neq \arg\max_{j} c_j(x_0)$. We call $r_a=\|x_a-x_0\| \le \alpha$ the perturbation radius of a successful attack.
\end{definition}
 \vspace{-0.3cm}
Ideally, the verification solution should provide the maximum radius $r$ of a safe norm ball as the verified safe radius, i.e., the black circle in Figure \ref{fig-1}. However, most sound verifiers can only calculate a lower bound of the maximum safe radius, i.e., a radius that is smaller than $r$, such as $r_1$ and $r_2$. 
Distinguishing from baseline methods, DeepAgn can estimate the maximum safe radius. 
 \vspace{-0.4cm}
\section{Lipschitz Analysis on Neural Networks}
 \vspace{-0.3cm}
\label{sec:lipschitz}
This section will theoretically prove that most neural networks, including recurrent neural networks, are Lipschitz continuous. We first introduce the definition of Lipschitz continuity.
 \vspace{-0.1cm}
\begin{definition}[Lipschitz Continuity~\cite{o2006metric}]\label{def_lip_con}
	Given two metric spaces $(X, d_X)$ and $(Y, d_Y)$, where $d_X$ and $d_Y$ are the metrics on the sets $X$ and $Y$ respectively, a function $f: X\rightarrow Y$ is called {\em Lipschitz continuous} if there exists a real constant $K\geq0$ such that, for all $x_1, x_2 \in X$:
		$d_Y(f(x_1), f(x_2)) \le K d_X(x_1, x_2)$.
	$K$ is called the {\em Lipschitz constant} of $f$. The smallest $K$ is called {\em the Best Lipschitz constant}, denoted as $K_{best}$.
\end{definition}
 \vspace{-0.8cm}
\subsection{Lipschitz Continuity of FNN}
 \vspace{-0.2cm}
Intuitively, a Lipschitz constant quantifies the changing rate of a function's output with respect to its input. Thus, if a neural network can be proved to be Lipschitz continuous, then Lipschitz continuity can potentially be utilized to bound the output of the neural network with respect to a given input perturbation. The authors in \cite{szegedy2013intriguing,ruan2018reachability} demonstrated that deep neural networks with convolutional, max-pooling layer and fully-connected layers with ReLU, Sigmoid activation function, Hyperbolic Tangent, and Softmax activation functions are Lipschitz continuous. According to the chain rule, 
the composition of Lipschitz continuous functions is still Lipschitz continuous. Thus we can conclude that a majority of deep feedforward neural networks are Lipschitz continuous.

 \vspace{-0.5cm}
\subsection{Lipschitz Analysis on Recurrent Neural Networks}
 \vspace{-0.2cm}
In this paper, we further prove that any recurrent neural network with finite input is Lipschitz continuous. Different from FNNs, RNNs contain feedback loops for processing sequential data, which can be unfolded into FNNs by eliminating loops~\cite{goodfellow2016}.  

\vspace{-0.5cm}
\begin{figure}[!htb]
\vspace{-0.3cm}
	\begin{minipage}{0.5\linewidth}
		\centering
		\includegraphics[width=0.65\linewidth]{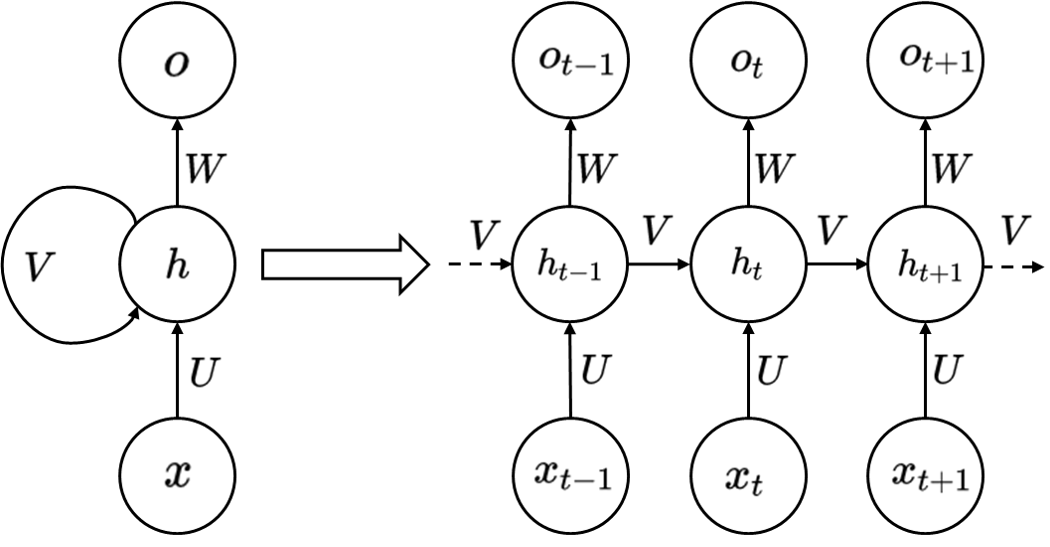}
	\end{minipage}
	\begin{minipage}{0.5\linewidth}
		\centering
		\includegraphics[width=0.65\linewidth]{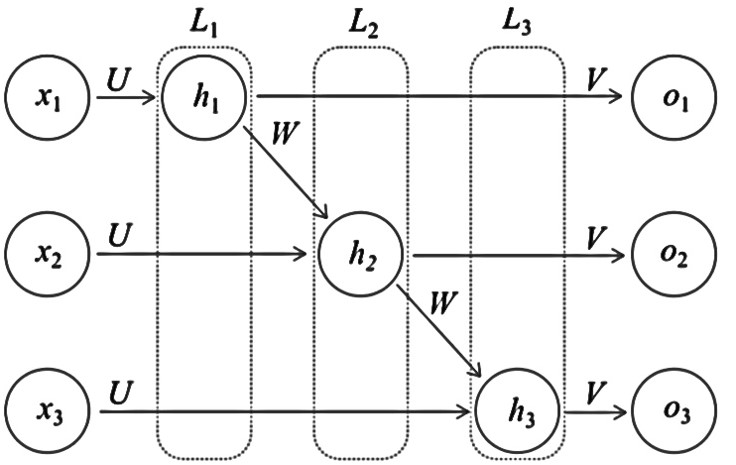}
		
	\end{minipage}
  \vspace{-0.4cm}
	\caption{(a) Unfolded recurrent neural network
	; 
	(b) A feedforward neural network by unfolding a RNN with input length 3: {\em The layers are denoted by $L_{i}$ for $1 \leq i \leq 3$. The node $h_2$ is located in the middle of layer $L_2$, taking $x_2$ and $h_1$. The rest of the $L_2$ are obtained by simply copying the information in $L_1$. }}
 \label{fig:unfolding of RNN}
 \vspace{-0.7cm}
\end{figure}
\begin{figure}[h]
\centering
\includegraphics[scale=0.36]{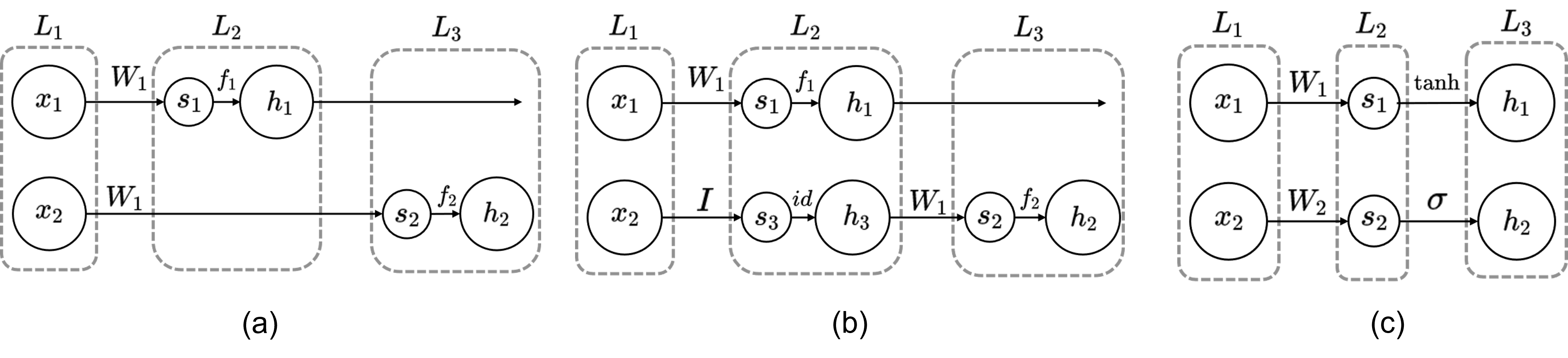}
\vspace{-0.4cm}
\caption{(a) Before we add dummy nodes to the intermediary layer: {\em $W_i$ is a weight matrix, and $f_i$ is an activation function. Initially, a connection from $x_2$ to $h_2$ crosses over the layer $L_2$.}
	~~(b) After we add dummy nodes: {\em we add nodes $s_3$ and $h_3$ to $L_2$, where $I$ denotes the identity matrix and $id$ denotes the identity function.}(c)Feed-forward layer with distinct activation functions: {\em  layer $L_2$ only performs a linear transformation (i.e. multiplication with weight matrices), and layer $L_3$ has the role of applying non-linear activation functions that contain two distinct activation functions: Hyperbolic Tangent and Sigmoid.}}
\label{fig:situations by unrollling}
\vspace{-1cm}
\end{figure}

Figure~\ref{fig:unfolding of RNN} illustrates such a process, by fixing the input size and
direct unrolling the RNNs, we can eliminate the loops and build an equivalent feed-forward neural network. The FNN however contains structures that do not appear in regular FNNs. 
They are time-delays between nodes in Figure \ref{fig:situations by unrollling} (a) and different activation functions in the same layer, see Figure \ref{fig:situations by unrollling} (c). 

For the time delay situation, we add dummy nodes to intermediary layers. These dummy nodes use the identity matrix for weight and use the identity function as an activation function, as illustrated in Figure~\ref{fig:situations by unrollling} (b). After the modification, the intermediary layer is equivalent to a regular FNN layer.

The time delay between nodes occurs even by simple structure RNNs, such as in Figure~\ref{fig:unfolding of RNN} (a), while the same layer with different activation functions appears only by unfolding complex RNNs. Figure \ref{fig:situations by unrollling} (c) demonstrates the layer with different activation functions after unrolling.
When the appeared different activation functions are Lipschitz continuous,  the layer is Lipschitz continuous based on the sub-multiplicative property in matrix norms.
See {\bf Appendix-A} for detailed proof.

\section{Reachability Analysis with Provable Guarantees}
\label{sec:approach}
\vspace{-0.3cm}



\subsection{Verification via Lipschitz Optimization}\label{sec:onedimensional}
\begin{figure}[H]
\vspace{-0.8cm}
	\centering
	\includegraphics[width=0.8\linewidth]{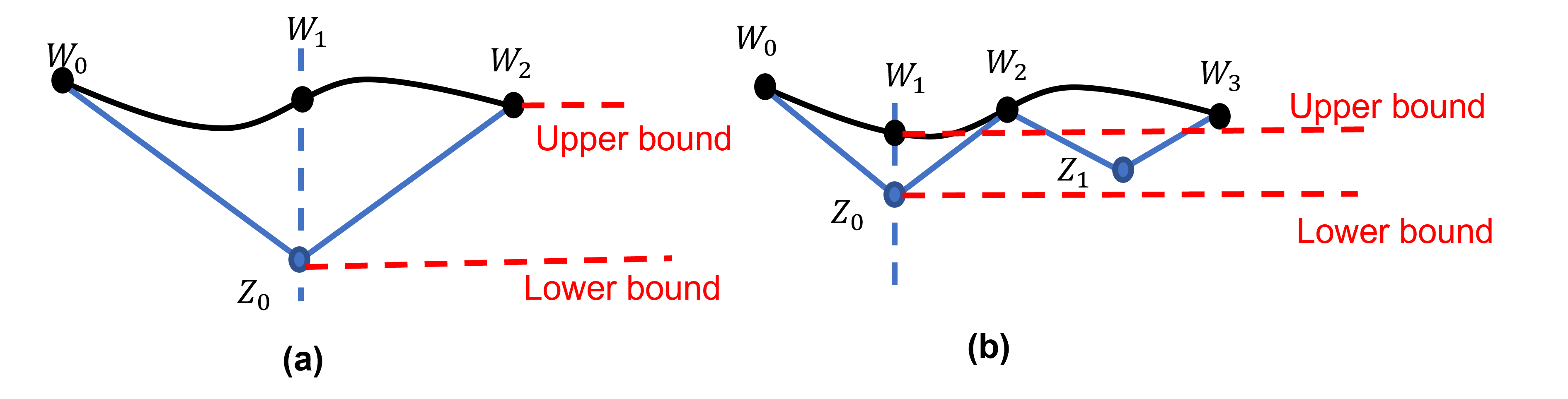}
 \vspace{-0.5cm}
	\caption{A lower-bound function designed via Lipschitz constant}
	\label{fig-6}
 \vspace{-0.5cm}
\end{figure}

\vspace{-0.1cm}
In the Lipschitz optimization~\cite{goldstein1977optimization} we asymptotically approach the global minimum. Practically, we execute a finite number of iterations by using an error tolerance $\epsilon$ to control the termination. As shown in Figure \ref{fig-6} (a), we first generate two straight lines with slope $K$ and $-K$, concreting a cross point $Z_0$. Since $Z_0$ is the minimal value of the generated piecewise-linear lower bound function (blue lines), we use the projected $W_1$ for the next iteration. In Figure \ref{fig-6} (b), new $W$ and $Z$ points are generated. In $i$-th iteration, the minimal value of $W$ is the upper bound $u_i$, and the minimal value of $Z$ is the lower bound $l_i$. Our approach constructs a sequence of lower and upper bounds, terminates the iteration whenever $|u_i-l_i|\leq\epsilon$,
%


\vspace{-0.3cm}
For the multi-dimensional optimization problem, we decompose it into a sequence of nested one-dimensional subproblems~\cite{gergel2016adaptive}.
Then the minima of those one-dimensional minimization subproblems are back-propagated into the original dimension, and the final global minimum is obtained with
\label{equ-16}
$\min\limits_{x \in [a_i,b_i]^n}~~\of(x) =\min\limits_{x_1\in [a_1,b_1]}... \min\limits_{x_n\in [a_n,b_n]} \of(x_1,...,x_n).
$
We define that for $1\leq k \leq n-1$,
	$\phi_k(x_1,...,x_k) = \min_{x_{k+1}\in [a_{k+1},b_{k+1}]} \phi_{k+1}(x_1,...,x_k, x_{k+1}) $
	and for $k=n$, $\phi_n(x_1,...,x_n) = \of(x_1,...,x_n).$
Thus we can conclude that  $ \min_{{x} \in [a_i,b_i]^n}~~\of({x}) = \min_{x_1\in [a_1,b_1]} \phi_1(x_1)$ which is actually a one-dimensional optimization problem.

We design a practical approach to dynamically update the current Lipschitz constant according to 
the previous iteration: 
$
K = \eta \max_{j = 1,...,i-1} \norm[\bigg]{\dfrac{\of(y_j) - \of(y_{j-1})}{y_j - y_{j-1}}}
$
where $\eta > 1$,  so that $ \lim_{i \to \infty} K > K_{best}.
$
~We use the Lipschitz optimisation to find the minimum and maximum function values of the neural network. With  binary search, we further estimate the maximum safe radius for target attack.
\vspace{-0.5cm}
\subsection{Global Convergence Analysis}
We first analyse the convergence for a one-dimensional case.
In the one dimensional case convergence exists under two conditions:  $\lim\limits_{i\to \infty}l_i = \min\limits_{x\in [a,b]} \of(x)$;  $\lim_{i\to \infty}(u_i - l_i) = 0$. It can be easily proved since  the lower bound sequence $\mathcal{L}_i$ is strictly monotonically increasing and bounded from above by $\min_{x\in [a,b]}\of(x)$. 


We use mathematical induction to prove convergence for the multi-dimension case.
%
The convergence conditions of the inductive step: if, for all $ {x}\in \mathbb{R}^k$, $\lim_{i\to \infty}l_i = \inf_{{x}\in [a,b]^k} \of({x})$ and $\lim_{i\to \infty}(u_i - l_i) = 0$ are satisfied, then, for all $ {x}\in \mathbb{R}^{k+1}$, $\lim_{i\to \infty}l_i = \inf_{{x}\in [a,b]^{k+1}} \of({x})$ and $\lim_{i\to \infty}(u_i - l_i) = 0$ hold.
\vspace{-0.2cm}
\begin{proof}\let\qed\relax(sketch)
	By the nested optimisation scheme, we have
$
	\min_{\mathbf{x} \in [a_i,b_i]^{k+1}}~~\of(\mathbf{x}) =\min_{x \in [a,b]}\Phi(x)
	, 
	\quad \Phi(x) = \min_{\mathbf{y} \in [a_i,b_i]^k} \of(x,\mathbf{y})
$.
	Since $\min_{\mathbf{y} \in [a_i,b_i]^k} \of(x,\mathbf{y})$ is bounded by an interval error $\epsilon_{\mathbf{y}}$, assuming $\Phi^*(x)$ is the accurate global minimum, then we have
	$\Phi^*(x)-\epsilon_{\mathbf{y}}\leq \Phi(x) \leq \Phi^*(x)+\epsilon_{\mathbf{y}}$
 $\Phi(x)$ is not accurate but bounded by $|\Phi(x) - \Phi^*(x)|\leq \epsilon_{\mathbf{y}}, \forall x \in [a,b]$, where $\Phi^*(x)$ is the accurate function evaluation.
	
	
	For the inaccurate evaluation case, we assume $\Phi_{min} = \min_{x\in [a,b]} \Phi(x)$, and its lower and bound sequences are, respectively, $\{l_0, ..., l_i\}$ and $\{u_0, ..., u_i\}$. The termination criteria for both cases are $|u^*_i-l^*_i|\leq \epsilon_x$ and $|u_i-l_i|\leq \epsilon_x$, and $\phi^*$ represents the ideal global minimum. Then we have $\phi^* - \epsilon_x \leq l_i$. Assuming that $l^*_i \in [x_k,x_{k+1}]$ and $x_k, x_{k+1}$ are adjacent evaluation points, then due to the fact that $ l^*_i = \inf_{x\in [a,b]} H(x;\mathcal{Y}_{i}) $ 
 and the search scheme, 
 we have $\phi^* - l_i \leq \epsilon_{\mathbf{y}} +\epsilon_x $. 
	Similarly, we can get $\phi^* + \epsilon_x \geq u^*_i = \inf_{y\in \mathcal{Y}_i } \Phi^*(y) \geq u_i -\epsilon_{\mathbf{y}}$
	so $u_i - \phi^* \leq \epsilon_x + \epsilon_{\mathbf{y}}$. By $\phi^* - l_i \leq \epsilon_{\mathbf{y}} +\epsilon_x$ and the termination criteria $u_i - l_i \leq \epsilon_x$, we have $l_i - \epsilon_{\mathbf{y}} \leq \phi^* \leq u_i + \epsilon_{\mathbf{y}}$, \ie~the accurate global minimum is also bounded. See more theoretical analysis of the global convergence in {\bf Appendix-B}. \qed
\end{proof} 
\vspace{-0.2cm}

\vspace{-0.4cm}
\section{Experiments}
\label{sec:experiment}
\vspace{-0.2cm}
\subsection{Performance Comparison with State-of-the-art Methods}
In this section, we compare DeepAgn with baseline methods. Their performance in feedforward neural networks and more details of the technique are demonstrated in {\bf Appendix-C}. Here, we mainly focus on the verification of RNN. 
We choose POPQORN~\cite{ko2019popqorn} as the baseline method since it can solve RNN verification problems analogously, i.e., calculating safe input bounds for given samples. Both methods were run on a PC with an i7-4770 CPU and 24 GB RAM. Table \ref{table1} demonstrates the verified safe radius of baseline $r_{b}$, DeepAgn $r$, and the radius of CW attack $r_a$. We fixed the number of hidden neurons and manipulated the input lengths in Table~\ref{table2} to compare the average safe radius and the time costs.
It can be seen that increasing the input length does not dramatically increase the time consumption of DeepAgn because it is independent of the models' architectures.
As in Table~\ref{table3}, we fixed the input length and employed RNNs and LSTMs with different numbers of hidden neurons.

\begin{table}[ht]\scriptsize
\centering
\caption{Average radius of attack and standard deviations~$(\cdot/\cdot)$ of Attack, DeepAgn, and POPQORN on MNIST}
\vspace{-0.3cm}
\label{table1}
\begin{tabular}{lccc}
\toprule[0.8pt]
 Model& ~~~~~~Attack ($r_a$)~~~~& \textbf{~~DeepAgn ($r$)~~~} & ~~POPQORN ($r_b$)\\ 
\midrule
 rnn 7\_64 & 0.8427/0.3723   & \textbf {0.5227/0.2157}   & 0.0198/0.014 \\
 rnn 4\_32 &  0.8424/0.4641   &  \textbf {0.6189/0.3231} & 0.0182/0.0201 \\
 lstm 4\_32 & 0.6211/ 0.4329 &  \textbf {0.3223/0.3563} &0.0081/0.0052 \\  
 lstm 7\_64 & 0.7126/ 0.3987   &  \textbf {0.4023/0.2112}   & 0.0194/0.0165 \\
\bottomrule
\end{tabular}
\vspace{-1cm}
\end{table}

\begin{table}[htb]
\setlength\tabcolsep{0.5pt}
\begin{minipage}[ct]{0.45\textwidth}\scriptsize
\centering
\caption{Average safe radius and time cost of DeepAgn and POPQORN on NN verification with different frame lengths}
\label{table2}
\begin{tabular}{lcccc} 
\toprule[0.8pt]
\multirow{2}{*}{Models}& \multicolumn{2}{c}{\textbf{DeepAgn}}&\multicolumn{2}{c}{POPQORN} \\ \cmidrule{2-5}
&\textbf{safe radius} &\textbf{time} &safe radius &time \\
\midrule
 rnn 4\_64&\textbf{0.1336} & 253.64s &0.0328 &1.31s \\ 
 rnn 14\_64&\textbf{ 0.3248} & 228.23s  & 0.2344 &11.73s \\ 
 rnn 28\_64&\textbf{0.3551}&\textbf{285.35s} & nan&nan  \\ 
 rnn 56\_64&\textbf{0.4369}&\textbf{314.1s} & nan&nan  \\ 
 lstm 4\_64 &\textbf{0.3195}&\textbf{250.99s} & 0.0004&307.93s\\
 lstm 14\_64&\textbf{0.3883}&\textbf{382s} & 0.0123&400.83s \\
 lstm 28\_64 &\textbf{0.6469}&\textbf{512.45s} & 0.0296&532.47s \\
 lstm 56\_64 &\textbf{0.6344}&\textbf{491.64s} & 0.0309&557.22s\\
\bottomrule
\end{tabular}
\end{minipage}
\hspace{1cm}
\begin{minipage}[ct]{0.43\textwidth}\scriptsize
\centering
\caption{Average safe radius and time cost of DeepAgn and POPQORN on NN verification with different hidden neurons}
\label{table3}
\begin{tabular}{lcccc} 
\toprule[0.8pt]
\multirow{2}{*}{Models}& \multicolumn{2}{c}{\textbf{DeepAgn}}&\multicolumn{2}{c}{POPQORN} \\ \cmidrule{2-5}
&\textbf{\makecell{safe\\radius}}&\textbf{time} &\makecell{safe\\radius}&time \\
\midrule
 rnn 7\_16&\textbf{0.5580}&~117.82s &~0.2038&~2.14s\\ 
 rnn 7\_32&\textbf{0.2371}&~175.92s& ~0.1340&~2.44s \\ 
 rnn 7\_128&\textbf{0.6633}&~240.59s & ~0.1052&~4.25s\\ 
 rnn 7\_256&\textbf{0.6656}&~187.83s & ~0.2038&~1.89s \\ 
 lstm 7\_16 &\textbf{0.3789}&\textbf{~175.11s} & ~0.0007&~243.60s\\
 lstm 7\_32 &\textbf{0.3461}&\textbf{~189.51s} & ~0.0015&~256.77s\\
 lstm 7\_128~~&\textbf{0.3625}&\textbf{~256.50s} & ~0.0050&~375.85s\\
 \bottomrule
 \end{tabular}
\end{minipage}
\end{table}
 \vspace{-0.5cm}
\subsection{Ablation Study}
\begin{figure}[h]
\vspace{-0.5cm}
  \centering
\includegraphics[width=0.8\textwidth]{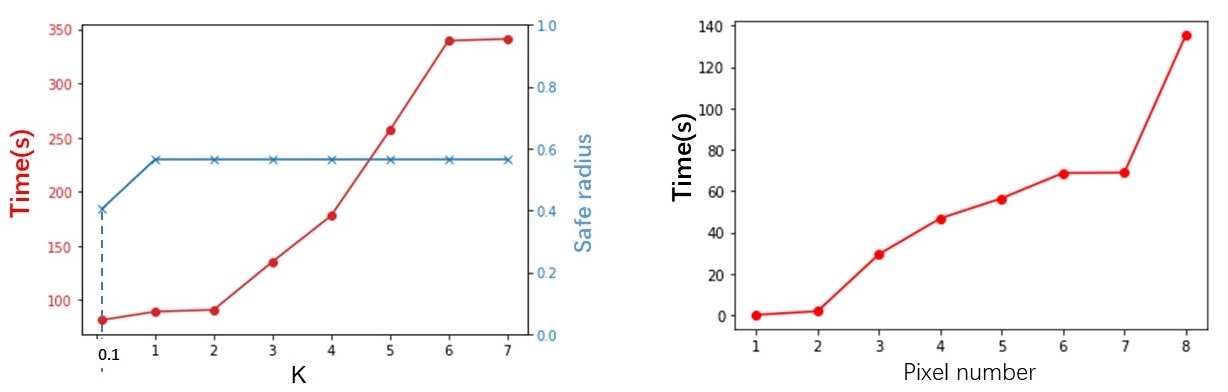}
\vspace{-0.5cm}
  \caption{Time cost and safe radius with  different $K$ and perturbed pixel numbers}
  \label{bound_k_n}
\vspace{-0.8cm}
\end{figure}

 \vspace{-0.2cm}
\label{ablation study}
In this section, we present an empirical analysis of the Lipschitz constant $K$ and the number of perturbed pixels, which both affect the precision of the results and the cost of time.
As shown in Figure~\ref{bound_k_n}, DeepAgn with $K=0.1$ gives a false safe radius, indicating that $0.1 $ is not a suitable choice. 
When the Lipschitz constant is larger than the minimal Lipschitz constant ($K \ge 1$), DeepAgn can always provide the exact maximum safe radius. 
However, with larger $K$, we need more iterations to achieve the convergence condition when solving the optimisation problem.
As for the number of perturbed pixels, we treat an n-pixel perturbation as an n-dimensional optimisation problem. Therefore, when the number of pixels increases, the evaluation time grows exponentially. 

 \vspace{-0.5cm}
\subsection{Case Study 1}
\begin{figure}[hb]
\vspace{-1cm}
  \centering
  \includegraphics[width=0.8\textwidth]{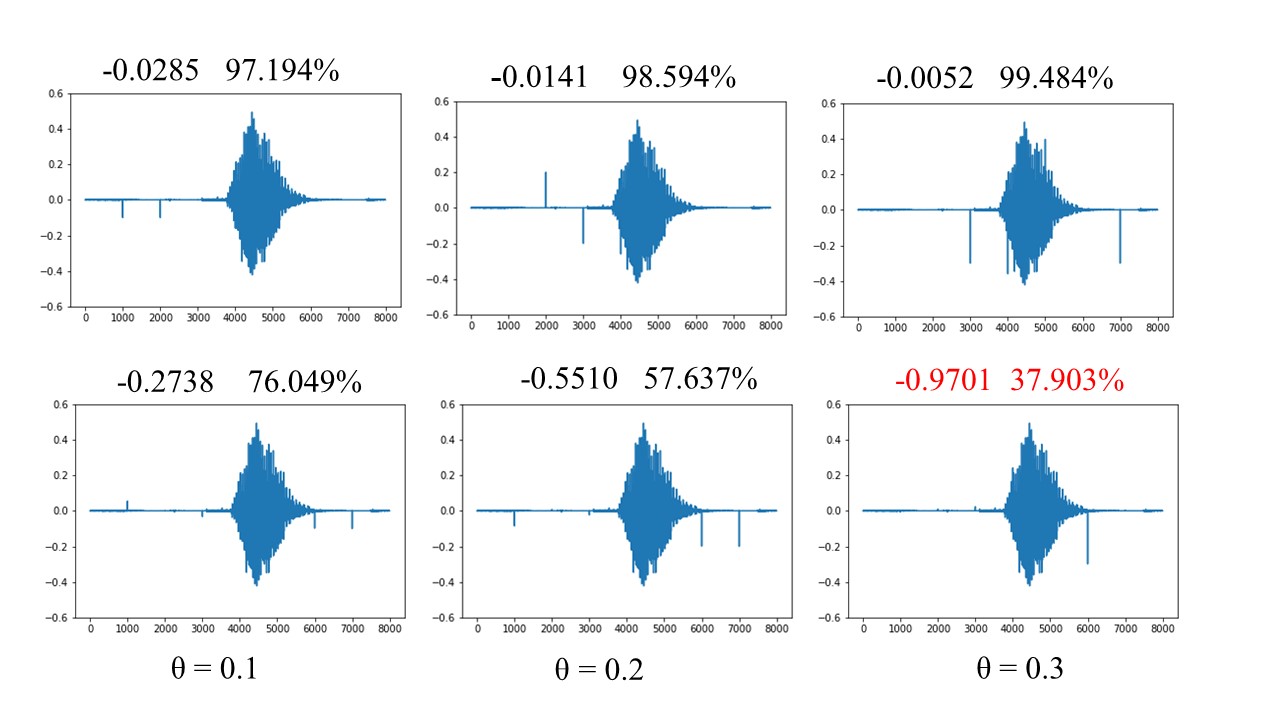}
  \vspace{-0.5cm}
  \caption{Input on lower and upper bounds of different perturbations. The first value indicates the logit output and the second shows the confidence value. For perturbation $\theta=0.1$ and $\theta=0.2$, the network remains safe. It is not safe for perturbation $\theta=0.3$. }
  \label{boundary2}
  \vspace{-0.7cm}
\end{figure}

\vspace{-0.2cm}
In this experiment, we use our method to verify a deep neural network in an audio classification task. 
The evaluated model is a deep CNN and is trained under the PyTorch framework.
The data set is adopted from~\cite{warden2018speech}, where each one-second raw audio is transformed into a sequence input with 8000 frames and classified into 35 categories. 
We perturb the input value of the frame $(1000,2000...,7000)$ and verify the network of different perturbation radii. 
For the deep CNN case, the baseline method has a lower verification accuracy, while DeepAgn can still provide the output ranges and the maximal safe radius. Figure~ \ref{boundary2} shows the boundary of the radio waveform with perturbation $\theta=0.1$, $\theta=0.2$ and  $\theta=0.3$. Their differences from the original audio are imperceptible to human ears. We performed a binary search and found the exact maximum safe radius $r=0.1591$.

\vspace{-0.5cm}
\begin{figure}[tb]
\vspace{-0.2cm}
  \centering
\includegraphics[width=0.6\textwidth]{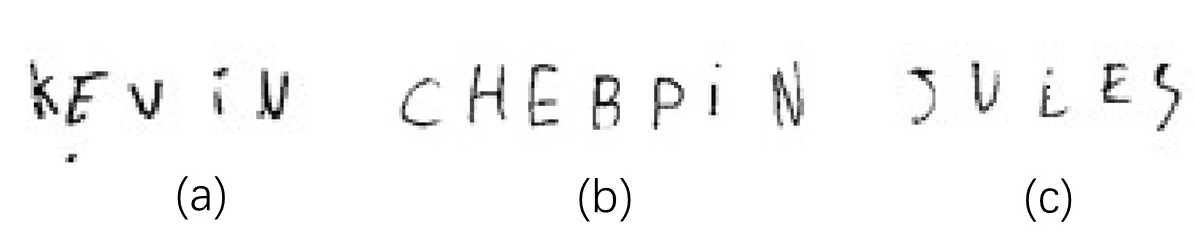}
  \vspace{-0.6cm}
  \caption{We perturb six pixels of the image to generate the ground-truth adversarial examples(a) $\theta= 0.638$, the first letter recognized as "K" with 48.174\% confidence, as "L" with 48.396\%,  the word  recognized as "IKEVIN"; (b) $\theta=0.0631$, confidence for fourth letter recognized as "R" 49.10\%,  "B" 49.12\%, recognized as "CHEBPIN" (c) $\theta=0.464$, the third letter as "L" 25.96\%, as "I" 25.97\%, recognized as "JUIES".}
  \label{boundary4}
  \vspace{-0.6cm}
 \end{figure}

\subsection{Case Study 2}
  \vspace{-0.2cm}
In this case study, we verify a hybrid neural network CRNN that contains convolutional layers and LSTM layers with CTC loss. The network converts characters from scanned documents into digital forms. As far as we know, there is no existing verification tool that can deal with this complex hybrid network. However, DeepAgn can analyze the output range of this CRNN and compute the maximum safe radius of a given input. In Figure~\ref{boundary4}, we present the maximum safe radius of the inputs and their associated ground-truth (or provably minimally-distorted) adversarial examples.

\vspace{-0.4cm}

\section{Conclusion}
\vspace{-0.2cm}
\label{sec:conclusion}
We design and implement a safety analysis tool for neural networks, computing reachability with provable guarantees. 
We demonstrate that it can be deployed in any network, including FNNs and RNNs regardless of the complex structure or activation function, as long as the network is Lipschitz continuous.  We envision that DeepAgn marks an important step towards practical and provably-guaranteed verification for DNNs. Future work includes using parallel computation and 
GPUs to improve its scalability on large-scale models trained on ImageNet, and generalising this method to other deep models such as deep reinforcement learning and transformers.



%

%
%
\vspace{-0.2cm}
\bibliographystyle{splncs04}
\bibliography{PAKDD_CR}

\end{document}